\documentclass[11pt]{article}

\usepackage[final]{acl}

\usepackage{times}
\usepackage{latexsym}
\usepackage[table]{xcolor}
\usepackage{colortbl}
\usepackage[T1]{fontenc}

\usepackage{enumitem}
\usepackage{amsmath}

\usepackage{microtype}

\usepackage{inconsolata}
\usepackage{graphicx}
\usepackage{subcaption}
\usepackage{booktabs}
\usepackage{amssymb}
\usepackage{multirow}   
\usepackage{tabularx}  
\usepackage{xcolor} 

\title{From Insight to Action: A Novel Framework for Interpretability-Guided Data Selection in Large Language Models}

\author{
  Ling Shi\textsuperscript{1}, \quad
  Xinwei Wu\textsuperscript{1,2}, \quad
  Xiaohu Zhao\textsuperscript{2},
  Hao Wang\textsuperscript{2},\quad
  Heng Liu\textsuperscript{2}, \quad \\ \bf
  Yangyang Liu\textsuperscript{2}, \quad
  Linlong Xu\textsuperscript{2}, \quad 
  Longyue Wang\textsuperscript{2}, \quad
  Deyi Xiong\textsuperscript{1}\thanks{~Corresponding author.}, \quad 
  Weihua Luo\textsuperscript{2}, \quad 
   \normalfont \\
  \textsuperscript{1}TJUNLP Lab, School of Computer Science and Technology, Tianjin University, China\quad \\
  \textsuperscript{2}Alibaba Group, China \\ 
   \texttt{\{shiling\_100, wuxw2021, dyxiong\}@tju.edu.cn} \\
}

\begin{document}
\maketitle

\begin{abstract}
While mechanistic interpretability tools like Sparse Autoencoders (SAEs) can uncover meaningful features within Large Language Models (LLMs), a critical gap remains in transforming these insights into practical actions for model optimization. 
We bridge this gap with the hypothesis that data selection guided by a model's internal task features is a effective training strategy.
Inspired by this, we propose Interpretability-Guided Data Selection (IGDS), a framework that first identifies these causal task features through frequency recall and interventional filtering, then selects ``Feature-Resonant Data'' that maximally activates task features for fine-tuning.
We validate IGDS on mathematical reasoning, summarization, and translation tasks within Gemma-2, LLaMA-3.1, and Qwen3 models.
Our experiments demonstrate exceptional data efficiency: on the Math task, IGDS surpasses full-dataset fine-tuning by a remarkable \textbf{17.4\%} on Gemma-2-2B while using only 50\% of the data, and outperforms established baselines focused on data quality and diversity.
Analysis confirms a strong positive correlation between feature amplification and task performance improvement.
IGDS thus provides a direct and effective framework to enhance LLMs by leveraging their internal mechanisms, validating our core hypothesis.
\end{abstract}

\section{Introduction}
\label{sec:introduction}
Large Language models (LLMs) have demonstrated increasingly superior performance across diverse downstream tasks~\citep{DBLP:journals/talip/LiuJSYX25, DBLP:journals/corr/abs-2502-20868, DBLP:conf/coling/ZhangX25, DBLP:conf/acl/0009YLJ00ZJCLSZ24,DBLP:conf/emnlp/PengSWZLLX25,shi2026sageserviceagentgraphguided}. 
Recent research in mechanistic interpretability has revealed that LLMs are not entirely black boxes; instead, they contain disentangled, human-understandable components~\citep{gaoscaling2024,arditi2024refusal}. 
Discoveries such as steering vectors for factual knowledge~\citep{ferrandoknow2024} and sparse features for cross-modal entities~\citep{lousae2025} have provided invaluable \textit{insights} into model mechanisms.
However, while these insights are powerful for analysis, a critical gap remains in translating them into practical \textit{actions} for model optimization~\citep{rai2024practical,sharkey2025open}.

\begin{figure}[t!]
    \centering
    \includegraphics[width=0.8\linewidth]{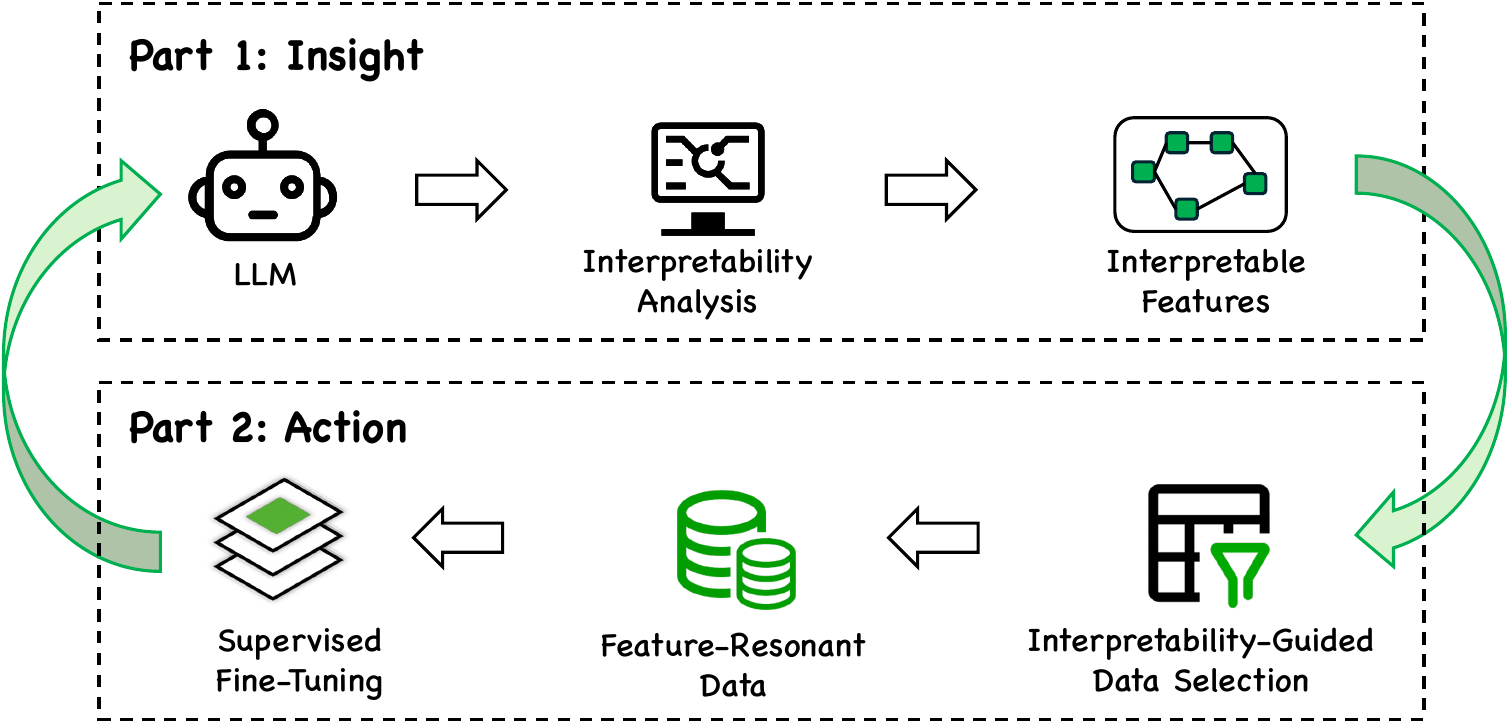} 

    \caption{Conceptual illustration of the IGDS paradigm. The diagram depicts the closed loop from internal insight to optimization action, showing how model features are leveraged to guide data selection.}

    \label{fig:intro_loop}
\end{figure}

We bridge this gap with the hypothesis that data selection guided by a model's internal, causally-validated task features is a highly effective training strategy.
To operationalize this, we propose \textbf{Interpretability-Guided Data Selection (IGDS)}, a framework that transforms interpretability insights into a tangible optimization pipeline.
This conceptual loop, which we term \textit{Insight2Action}, is visualized in Figure~\ref{fig:intro_loop}.
Our core idea is to first identify the model's beneficial internal mechanisms and then select data that maximally activates them, which is called ``Feature-Resonant Data'', to reinforce these mechanisms through fine-tuning.

\begin{figure*}[t!]
    \centering
    \includegraphics[width=\textwidth]{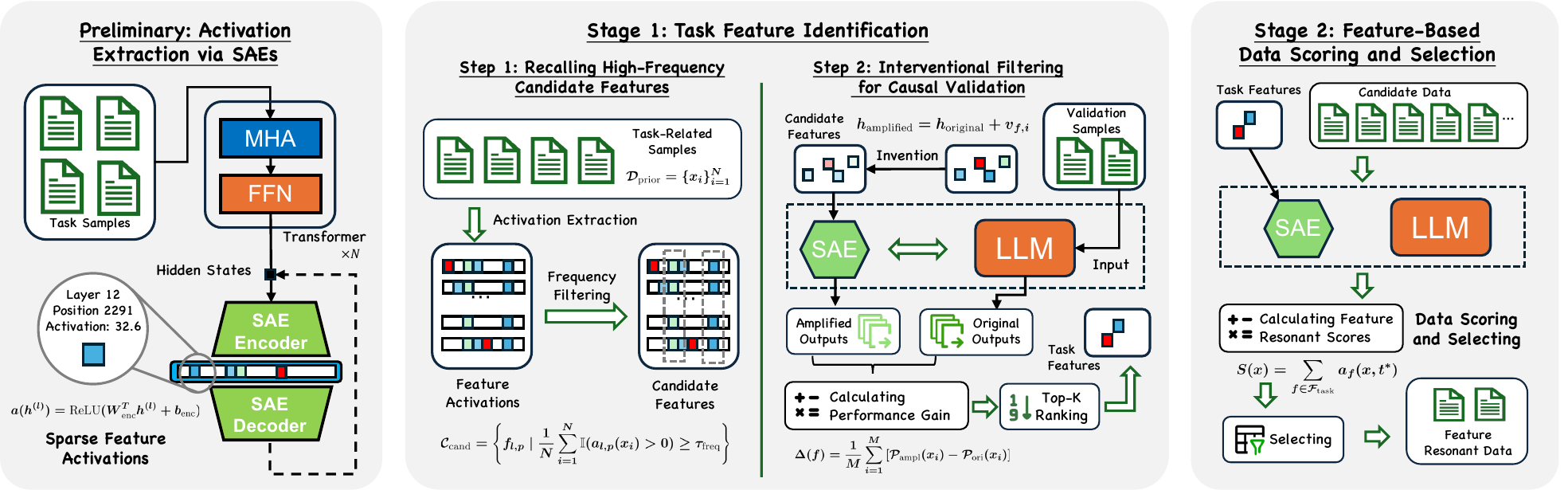}
    \caption{An overview of our Interpretability-Guided Data Selection (IGDS) framework.}
    \label{fig:overview}
\end{figure*}

Specifically, IGDS operationalizes this vision through a two-stage process. The first stage, \textbf{Task Feature Identification}, moves from a broad \textit{high-frequency recall} of candidate features using an SAE to a rigorous \textit{causal intervention filtering} step, which isolates the potent subset directly impacting task performance. The second stage, \textbf{Feature-Based Data Scoring}, then uses this validated set of causal features to construct a highly effective utility function, selecting data that ``resonates'' with and reinforces the model's internal problem-solving structure.

We validate IGDS on mathematical reasoning, summarization, and translation tasks across Gemma-2, LLaMA-3.1, and Qwen3 models. 
Our experiments show that our method achieves the highest data efficiency across all settings.
Notably, on the Math task, IGDS surpasses full-dataset fine-tuning by a remarkable \textbf{17.4\%} on the Gemma-2-2B model while using only 50\% of the data, and consistently outperforms established baselines focused on data quality and diversity. 
Furthermore, our analysis confirms a strong positive correlation between the targeted amplification of these internal features and the improvements in downstream task performance, providing strong mechanistic evidence for our method's success.

In summary, our key contributions are:
\begin{itemize}
    \item We propose a novel optimization strategy that leverages a model's causally validated internal features to guide data selection, offering a direct path to efficient capability enhancement.

    \item We introduce IGDS, a general and practical framework that transforms descriptive interpretability insights into a prescriptive pipeline for identifying and selecting high-utility training data, effectively closing the loop from analysis to optimization.

    \item We provide extensive empirical validation across multiple models and tasks, demonstrating that IGDS surpasses both competitive baselines and full-dataset fine-tuning while utilizing only a fraction of the data.

\end{itemize}

\section{Related Work}

\paragraph{Mechanistic Interpretability} Mechanistic interpretability (MI) aims to reverse-engineer Large Language Models (LLMs) to uncover the causal mechanisms behind their behaviors~\citep{dunefsky2024transcoders,sharkey2025open}.
Recent breakthroughs have revealed that fine-grained semantic information is encoded within model representations, from neurons corresponding to specific concepts~\citep{niudoes2024,fang2024towards}, to steering vectors that control high-level attributes like factuality and safety~\citep{ferrandoknow2024,yi2025nlsr}, and sparse features that relevant to downstream tasks~\citep{lusparse,shu2025survey}.
However, most MI research currently stops at the explanatory level, further exploration into how these insights can proactively build better models is often neglected~\citep{rai2024practical,sharkey2025open}.
While prior research has begun to explore applications, such as controlling model behavior via inference-time interventions~\citep{wu2024mitigating,ghosh2025simple} or refining reward models specifically for safety alignment~\citep{zhang2025interpretable,liu2025sparserm}, a general pipeline for leveraging these insights to guide the broader training process remains absent. This work aims to bridge this gap, establishing a framework that turns descriptive interpretability findings into a prescriptive guide for model optimization.

\paragraph{Sparse Autoencoders} Sparse Autoencoders (SAEs) have emerged as a pivotal tool in MI for addressing the superposition hypothesis~\citep{sharkey2025open}. This premise suggests that densely parameterized models frequently superimpose multiple independent semantic signals onto individual neuronal activations. Such polysemanticity obscures interpretability at both the single-unit and embedding-vector scales~\citep{elhage2022toy,gurnee2023finding}. By untangling these entangled representations into a sparse basis of semantically coherent dimensions, SAEs markedly improve architectural transparency~\citep{cunningham2023sparse,gao2024scaling}. Current applications of SAEs typically focus on identifying task-specific features to enable precise, post-hoc interventions on model behavior~\citep{templeton2024scaling,farrell2024applying, DBLP:conf/aaai/WuLZRXWXLZ26}.
While effective for changing outputs, such interventional approaches can suffer from high latency and instability in practical applications.
Departing from this paradigm, we leverage SAE-identified features to guide the data selection process, establishing a new pathway from internal insight to optimization action.

\paragraph{Data Selection} Training data remains a critical determinant of model efficacy~\citep{DBLP:journals/corr/abs-2408-06273, anonymous2026from,DBLP:journals/corr/abs-2507-09205}.  The paradigm for post-training data selection has shifted from prioritizing quantity to emphasizing quality~\citep{zhou2023lima}, giving rise to automated methods that score data based on \textbf{data quality} using external models~\citep{chen2023alpagasus} or self-assessed metrics~\citep{li2023quantity,xia2024less}.
Beyond quality, \textbf{data diversity} is recognized as crucial for model robustness, with frameworks like \textit{ZIP} proposed to balance this trade-off~\citep{bukharin2024data,yin2024entropy}.
However, the efficacy of these methods is scrutinized at scale, as recent studies suggest they often yield negligible gains over simple random selection~\citep{xia2024rethinking}. Departing from prior approaches that rely on external signals and treat the model as a black box, we posit that a direct and more potent signal for data utility resides within the model itself.

\section{Methodology}
\label{sec:methodology}

In this section, we detail the proposed Interpretability-Guided Data Selection (IGDS) framework, which operationalizes interpretability insights into a rigorous data selection strategy. As illustrated in Figure~\ref{fig:overview}, the framework proceeds in two main stages: (1) \textbf{Task Feature Identification}, where we isolate features that are not merely correlated with, but causally linked to task performance; and (2) \textbf{Feature-Based Data Scoring}, where we leverage these validated features to quantify data utility and curate high-quality subsets for supervised fine-tuning.

\subsection{Preliminaries: Activation Extraction via SAEs}
\label{subsec:preliminaries}

Our framework leverages Sparse Autoencoders (SAEs) to transform compact neural activations into a sparse, expansive representation where individual dimensions carry distinct semantic meaning.
Given an input sequence, we first extract the hidden state $\boldsymbol{h}^{(l)} \in \mathbb{R}^{d_{\text{model}}}$ from a specific layer $l$ via a forward pass.
The SAE then projects this dense representation into a sparse feature activation vector $\boldsymbol{a}(\boldsymbol{h}^{(l)}) \in \mathbb{R}^{d_{\text{sae}}}$ using an encoder parameterized by the weight matrix $\boldsymbol{W}_{\text{enc}}$:
\begin{equation}
    \boldsymbol{a}(\boldsymbol{h}^{(l)}) = \text{ReLU}(\boldsymbol{W}_{\text{enc}}^T \boldsymbol{h}^{(l)} + \boldsymbol{b}_{\text{enc}}),
    \label{eq:sae_encoder}
\end{equation}
where $d_{\text{sae}} \gg d_{\text{model}}$.
Each dimension $p$ in this vector corresponds to a feature $f_{l,p}$, with its value indicating the feature's activation magnitude. Our goal is to navigate this vast feature space and pinpoint the precise subset of features that exert a causal influence on the target task.

\subsection{Stage 1: Task Feature Identification}
\label{subsec:feature_identification}

To identify these features, we employ a coarse-to-fine filtering strategy that distills the vast feature space through two sequential steps: high-frequency recalling and interventional filtering. 
\paragraph{Recalling High-Frequency Candidate Features}
A feature fundamental to a specific task is expected to activate consistently during task execution~\citep{geiger2024finding}. Based on this principle, we begin by identifying candidate features that exhibit a strong correlation with the target task.
Utilizing a small corpus of $N$ prior task-related samples, denoted as $\mathcal{D}_{\text{prior}} = \{x_i\}_{i=1}^N$, we monitor feature activations at the critical token position (e.g., the final token of the prompt), with specific positions for each task listed in Appendix~\ref{tab:prompts}. 
Let $a_{l,p}(x_i)$ denote the activation magnitude of feature $f_{l,p}$ for sample $x_i$ at this step.

A feature $f_{l,p}$ is selected as a candidate if its activation frequency, defined as the proportion of samples in $\mathcal{D}_{\text{prior}}$ where the feature is active, exceeds a predefined threshold $\tau_{\text{freq}}$ (e.g., 80\%).
Formally, we define the set of candidate features, $\mathcal{C}_{\text{cand}}$, as follows:
\begin{equation}
    \mathcal{C}_{\text{cand}} = \left\{ f_{l,p} \mid \frac{1}{N} \sum_{i=1}^{N} \mathbb{I}(a_{l,p}(x_i) > 0) \geq \tau_{\text{freq}} \right\},
    \label{eq:recall_features}
\end{equation}
where $\mathbb{I}(\cdot)$ is the indicator function, returning 1 if the condition holds and 0 otherwise.
This initial step effectively filters out the vast majority of irrelevant features, yielding a manageable candidate set for subsequent rigorous validation.

\paragraph{Interventional Filtering for Causal Validation}

High activation frequency implies correlation, but not necessarily causation.
A feature that consistently activates on task data might merely capture general linguistic patterns rather than task-specific mechanisms~\citep{deng2025sparse}. To isolate features with genuine causal efficacy, we perform a rigorous filtering step via targeted interventions on a small validation set of $M$ samples, $\mathcal{D}_{\text{val}} = \{x_i\}_{i=1}^M$.

For each candidate feature $f \in \mathcal{C}_{\text{cand}}$, we quantify its causal impact by measuring how {amplifying} its activation affects the model's performance.
This is achieved by adding the feature's influence vector to the model's residual stream and evaluating the resulting change in task-specific performance. 
Specifically, for each validation sample $x_i$, we compute the feature's influence vector $\boldsymbol{v}_{f,i}$ as the product of its current activation and its corresponding SAE decoder weight:
\begin{equation}
    \boldsymbol{v}_{f,i} = a(h_i) \cdot \boldsymbol{W}_{\text{dec}, f},
    \label{eq:feature_vector_revised}
\end{equation}
where $a(h_i)$ denotes the scalar activation of feature $f$ on the hidden state $h_i$, and $\boldsymbol{W}_{\text{dec}, f}$ represents the decoder weight vector for feature $f$.
We then generate two outputs: one from the original model, and one from an ``amplified'' counterpart where the feature's influence vector is directly {added} to the residual stream:
\begin{equation}
    \boldsymbol{h}_{\text{ampl}} = \boldsymbol{h}_{\text{ori}} + \boldsymbol{v}_{f,i},
    \label{eq:amplification}
\end{equation}
Let $\mathcal{P}_{\text{ori}}(x_i)$ and $\mathcal{P}_{\text{ampl}}(x_i)$ denote the task performance scores (e.g., COMET for translation) for the original and amplified outputs, respectively.
The causal impact of feature $f$ is defined as the average performance gain across the validation set:
\begin{equation}
    \Delta(f) = \frac{1}{M}\sum_{i=1}^{M} \left[ \mathcal{P}_{\text{ampl}}(x_i) - \mathcal{P}_{\text{ori}}(x_i) \right],
    \label{eq:perf_gain_revised}
\end{equation}
A significantly positive $\Delta(f)$ indicates that amplifying the feature consistently enhances the model's task capability, validating the feature as a positive causal driver. Finally, we rank all candidates by $\Delta(f)$ and select the top-$K$ features to form the final validated set of {task features}, $\mathcal{F}_{\text{task}}$. This interventional filtering ensures the retention of features that are demonstrably beneficial for the target task.

\subsection{Stage 2: Feature-Based Data Scoring and Selection}
\label{subsec:scoring}

With the causally-validated set of task-relevant features $\mathcal{F}_{\text{task}}$ identified, we leverage this subset as an intrinsic lens to quantify the utility of each candidate data point from a large pool $\mathcal{D}_{\text{pool}}$. 
Our central hypothesis posits that the most valuable data for fine-tuning are those that maximally activate the model's internal causal mechanisms associated with the task.

To operationalize this, we introduce the \textbf{Feature-Resonant Score (FRS)}, denoted as $S(x)$. 
This score is computed for each data point $x$ by aggregating the activation magnitudes of all task features at the same critical token position, $t^*$, consistent with the identification phase in Stage 1.
Formally, the score is defined as:
\begin{equation}
    S(x) = \sum_{f \in \mathcal{F}_{\text{task}}} a_f(x, t^*),
    \label{eq:frs_score}
\end{equation}
where $a_f(x, t^*)$ represents the activation of task feature $f$ at position $t^*$ for input $x$. By design, this formulation directly prioritizes data that elicits a strong, collective response from the specific features underpinning the desired capability.

Finally, we rank all data points in $\mathcal{D}_{\text{pool}}$ by their FRS and select a subset based on a predefined ratio. 
This process yields our final training dataset, $\mathcal{D}_{\text{SFT}}$, a high-potency subset of the original corpus, densely packed with targeted, task-relevant signals.

\begin{table*}[t!]
\centering
\caption{Results of the task feature identification. For each model and task, we report the proportion of high-frequency candidates in basis points (\textit{Recalled (\textpertenthousand)}, where 100\textpertenthousand = 1\%), the specific feature with the top-1 positive impact (\textit{Feature}), and its corresponding performance gain (\textit{$\Delta$}).}
\label{tab:feature_id_results}
\resizebox{\linewidth}{!}{%
\begin{tabular}{@{}l ccc ccc ccc@{}}
\toprule
& \multicolumn{3}{c}{Math} & \multicolumn{3}{c}{Summarization} & \multicolumn{3}{c}{Translation} \\
\cmidrule(lr){2-4} \cmidrule(lr){5-7} \cmidrule(lr){8-10}
Model & Recalled (\textpertenthousand) & Feature & $\Delta$(ACC) & Recalled (\textpertenthousand) & Feature & $\Delta$(ROUGE-1) & Recalled (\textpertenthousand) & Feature & $\Delta$(COMET) \\
\midrule
Gemma-2-2B-it & 9.20 & \texttt{l14\_p11575} & +12 & 8.89 & \texttt{l25\_p3017} & +0.025 & 4.82 & \texttt{l11\_p892} & +1.12 \\
LLaMA-3.1-8B-it & 2.76 & \texttt{l19\_p16897} & +1.5 & 3.40 & \texttt{l31\_p15962} & +0.022 & 4.91 & \texttt{l8\_p2083} & +4.98 \\
Qwen3-8B & 1.33 & \texttt{l16\_p36564} & +1.8 & 0.35 & \texttt{l18\_p61304} & +0.018 & 2.94 & \texttt{l12\_p43296} & +8.34 \\
\bottomrule
\end{tabular}%
}
\end{table*}

\section{Experiments}
\label{sec:experiments}


To validate the effectiveness and robustness of our Interpretability-Guided Data Selection (IGDS) framework, we conducted a comprehensive set of experiments across diverse tasks, models, and competitive baselines.

\subsection{Tasks and Evaluation}
\label{subsec:tasks_datasets_metrics}


For each task, we strictly enforced data separation to prevent information leakage. Specifically, we distinctively defined three subsets: the task-related set for feature identification (Stage 1), a large candidate pool for selection (Stage 2), and a held-out test set for final evaluation. Detailed prompt templates for these stages are provided in Appendix~\ref{app:prompt_templates}. The specific setup for each task is detailed below:

\paragraph{Mathematical Reasoning} 
The selection pool ($\mathcal{D}_{\text{pool}}$) comprises 93.7K samples from the \texttt{OpenR1-Math-220k} dataset. For feature identification, we utilized the training set of \texttt{gsm8k}~\citep{zeng2023challenge}. Final model performance was evaluated on the \texttt{MATH-500} benchmark.
\paragraph{Summarization} 
We utilized the \texttt{DialogSum} dataset~\citep{chen2021dialogsum}. The official training split (12.5K samples) served as the selection pool ($\mathcal{D}_{\text{pool}}$), while the validation split (500 samples) was employed for feature identification. Performance was measured on the official test split (1.5k samples).

\paragraph{Machine Translation} 
The selection pool ($\mathcal{D}_{\text{pool}}$) consists of 10K English-to-Chinese pairs randomly sampled from the \texttt{WMT24} dataset~\citep{kocmi2024findings}. From the same source, a separate set of 500 samples was randomly held out to serve as the set for feature identification. Evaluation was conducted on the \texttt{WMT24++} test set~\citep{deutsch2025wmt24++} (997 samples).

\paragraph{Evaluation Metrics}
We report performance using standard, task-specific metrics. For Mathematical Reasoning, we employ \texttt{pass@8} accuracy. For Summarization, we report \texttt{ROUGE-1} scores in the main text, with full \texttt{ROUGE-1/2/L} results provided in the Appendix~\ref{full-rouge}. For Machine Translation, we utilize the \texttt{COMET} score \cite{rei2020comet}. %

\subsection{Models and SAEs}
\label{subsec:models_saes}

Our experiments encompass three model families: \texttt{Gemma-2}~\citep{team2024gemma}, \texttt{LLaMA-3.1}~\citep{dubey2024llama}, and \texttt{Qwen3}~\citep{yang2025qwen3}. We adopted a strategic two-stage setup: task features were identified using the publicly available {instruction-tuned} versions, as these models possess the requisite task awareness for meaningful feature discovery~\citep{xia2024less}. The subsequent fine-tuning was then performed on the corresponding {base models} to cleanly evaluate the impact of the selected data. This setup ensures that any observed performance gains are attributed directly to the quality of the selected data, rather than confounding factors from the initial instruction tuning. 

For feature extraction, we utilized SAEs specific to models. For \texttt{Gemma-2} and \texttt{LLaMA-3.1}, we leveraged publicly available pre-trained SAEs~\citep{lieberum2024gemma,he2024llamascope}. In the absence of publicly available SAEs for \texttt{Qwen3}, we trained a custom instance using the JumpReLU architecture~\citep{he2024llamascope}. Full training details for this custom SAE are provided in Appendix~\ref{app:sae_training}.

\subsection{Baselines}
\label{subsec:baselines}
To comprehensively evaluate the IGDS framework, we benchmarked it against a spectrum of data selection baselines and standard controls. We compared IGDS against three distinct selection strategies: (1) \textbf{Quality-based} methods, including \textit{IFD}~\citep{li2023quantity} targeting instruction-following difficulty, and \textit{Loss}, a perplexity filter prioritization samples with the lowest Cross-Entropy; (2) A \textbf{Diversity-based} approach, \textit{ZIP}~\citep{yin2024entropy}, designed to maximize semantic coverage; and (3) \textit{Random} selection, which serves as a robust, data-agnostic baseline to validate the necessity of intelligent selection metrics. To contextualize performance, we establish two standard controls: \textit{Original}, evaluating the zero-shot capabilities of the base model without SFT, and \textit{Full}, utilizing the entire training pool as a reference for data efficiency.

\begin{table*}[t!]
\centering
\caption{Performance comparison across three tasks: Math (pass@8), Summarization (ROUGE-1), and Translation (COMET). The best result among all fine-tuning methods for each model-task pair is in \textbf{bold}. The subscript denotes the gap relative to the performance of full SFT.}
\label{tab:main_results}

\resizebox{\textwidth}{!}{%

\begin{tabular}{@{}llcc ccccc@{}}
\toprule
& & \multicolumn{2}{c}{Standard Controls} & \multicolumn{5}{c}{Data Selection Baselines} \\
\cmidrule(lr){3-4} \cmidrule(lr){5-9}

Task & Model & Original & Full & Random & Loss & IFD & ZIP & \textbf{IGDS} \\
\midrule
\multirow{3}{*}{Math} 
&Gemma-2-2B & $11.2$ & $32.2$ & $29.2_{-9.3\%}$ & $24.6_{-23.6\%}$ &$30.0_{-6.8\%}$ & $29.4_{-8.7\%}$ & $\mathbf{37.8_{+17.4\%}}$ \\
&Llama-3.1-8B & $26.6$ & $45.0$ & $40.8_{-9.3\%}$ & $37.2_{-17.3\%}$ & $36.8_{-18.2\%}$ & $35.4_{-21.3\%}$ & $\mathbf{45.8_{+1.8\%}}$ \\
&Qwen3-8B-Base & $55.7$ & $60.8$ & $58.7_{-3.5\%}$ & $59.4_{-2.3\%}$ & $60.5_{-0.5\%}$ & $60.3_{-0.8\%}$ & $\mathbf{61.1_{+0.4\%}}$ \\
\midrule
\multirow{3}{*}{Sum} 
& Gemma-2-2B & $0.220$ & $0.450$ & $0.439_{-2.4\%}$ & $0.448_{-0.6\%}$ & $0.440_{-2.2\%}$ & $0.449_{-1.2\%}$ & $\mathbf{0.452_{+0.4\%}}$ \\
&Llama-3.1-8B & $0.011$ & $0.260$ & $0.254_{-2.3\%}$ & $0.254_{-2.3\%}$ & $0.258_{-0.8\%}$ & $0.245_{-5.8\%}$ & $\mathbf{0.261_{+0.4\%}}$ \\
&Qwen3-8B-Base & $0.312$ & $0.489$ & $0.482_{-1.4\%}$ & $0.474_{-3.1\%}$ & $0.488_{-0.2\%}$ & $0.485_{-0.8\%}$ & $\mathbf{0.490_{+0.2\%}}$ \\

\midrule
\multirow{3}{*}{Trans} 
& Gemma-2-2B & $26.92$ & $65.93$ & $62.92_{-4.6\%}$ & $55.31_{-16.1\%}$ & $64.19_{-2.6\%}$ & $63.37_{-3.9\%}$ & $\mathbf{68.14_{+3.4\%}}$ \\
&Llama-3.1-8B & $31.11$ & $77.68$ & $70.78_{-8.9\%}$ & $72.29_{-6.9\%}$ & $76.82_{-1.1\%}$ & $74.49_{-4.1\%}$ & $\mathbf{78.45_{+1.0\%}}$ \\
&Qwen3-8B-Base & $81.30$ & $82.71$ & $82.61_{-0.1\%}$ & $82.66_{-0.1\%}$ & $82.69_{-0.0\%}$ & $82.57_{-0.2\%}$ & $\mathbf{83.07_{+0.4\%}}$ \\
\bottomrule
\end{tabular}%
}
\end{table*}

\subsection{Main Results}
\label{sec:main_results}

\paragraph{Results of Task-Specific Features Identification} We initiate our analysis by examining the efficacy of the feature identification stage, with key statistics summarized in Table~\ref{tab:feature_id_results}. The process begins by applying a frequency filter to the vast search space comprising millions of potential SAE features. This initial step proves highly selective, drastically reducing the vast search space. As shown in the \textit{Recalled} column, high-frequency candidates represent a minute fraction of the total, often amounting to just a few basis points (\textpertenthousand).

From this condensed pool, our causal validation step consistently identifies task features whose amplification yields a positive performance impact on the validation set.
The magnitude of this impact, reported in the \textit{$\Delta$} column, is often substantial.
We observe particularly striking efficacy in specific cases: for instance, \texttt{Gemma-2-2B} achieves a performance increase of 12 points on the Math task, while \texttt{Qwen3-8B} sees a massive gain of +8.34 on Translation.
The consistent discovery of such high-impact features across diverse models and tasks provides compelling evidence that our framework reliably identifies features that are not merely correlated with, but causally instrumental to, the model's task-solving capabilities.

\begin{figure}[t!]
    \centering
    \includegraphics[width=0.95\linewidth]{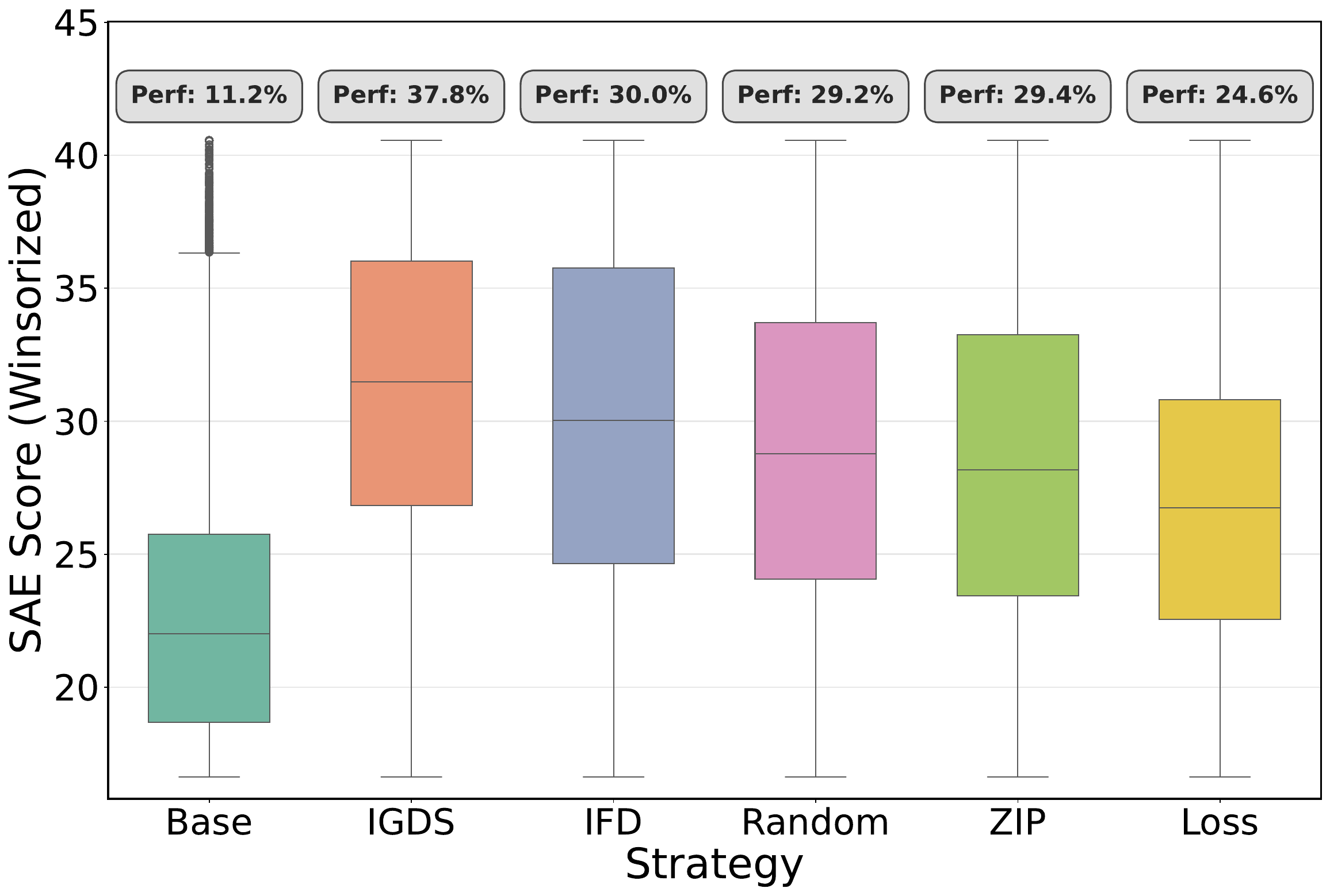}
    \caption{Correlation between feature activation and task performance on the Math task for Gemma-2-2B. }
    \label{fig:activation_vs_perf}
\end{figure}

\paragraph{Performance of IGDS framework} Utilizing the identified features, we selected the top 50\% of the candidate pool to fine-tune base models. Table~\ref{tab:main_results} provides a comprehensive comparison against all baselines and controls.

The results show IGDS consistently and substantially outperforms all other data selection methods across every tested model and task.
Furthermore, our method surpasses the performance of full-data fine-tuning in multiple scenarios.
As indicated by the positive subscripts in Table~\ref{tab:main_results}, IGDS even achieves a striking +17.4\% relative gain for Gemma-2-2B on Math. 
This phenomenon strongly validates our core hypothesis: selecting data through the lens of the model's own causally-validated mechanisms is a effective strategy for targeted model improvement.

\subsection{Validating the Role of Task Features}

To corroborate the link between our identified features and downstream task proficiency, we analyzed the post-fine-tuning activation distributions of a key task feature. Figure~\ref{fig:activation_vs_perf} shows the results for the top-ranked Math feature, \texttt{l14\_p11575}, in the \texttt{Gemma-2-2B} model. We plot its activation distribution across the training set for the base model and for models fine-tuned with various data selection strategies, alongside the final task performance. 

Two key observations emerge from our analysis. First, even baseline strategies that do not explicitly target this feature (e.g., \textit{Random}, \textit{IFD}) implicitly enhance its activation levels compared to the base model. This suggests that the feature is intrinsically aligned with the underlying optimization mechanics of the Math task. Second, and most crucially, there is a strong positive correlation between elevated activation magnitude and optimized task performance. The IGDS-trained model not only exhibits the highest median feature activation but also achieves the superior performance score of 37.8. Furthermore, a consistent trend is evident across all methods: the hierarchy of median feature activations closely mirrors the ranking of final performance. Collectively, these findings provide compelling evidence that the features identified by our framework are causally instrumental to the model's task-solving capabilities. 

\section{Robustness Analysis}
Beyond standard performance benchmarks, we further investigate the practical viability of the IGDS framework. This section demonstrates the method's consistent superiority across varying data budgets, validates the contribution of its core components via ablation studies, and confirms its computational efficiency, solidifying its standing as a practical and reliable solution.

\subsection{Effect of Different Sampling Ratios}

\begin{figure}[t!]
    \centering
    \includegraphics[width=0.95\linewidth]{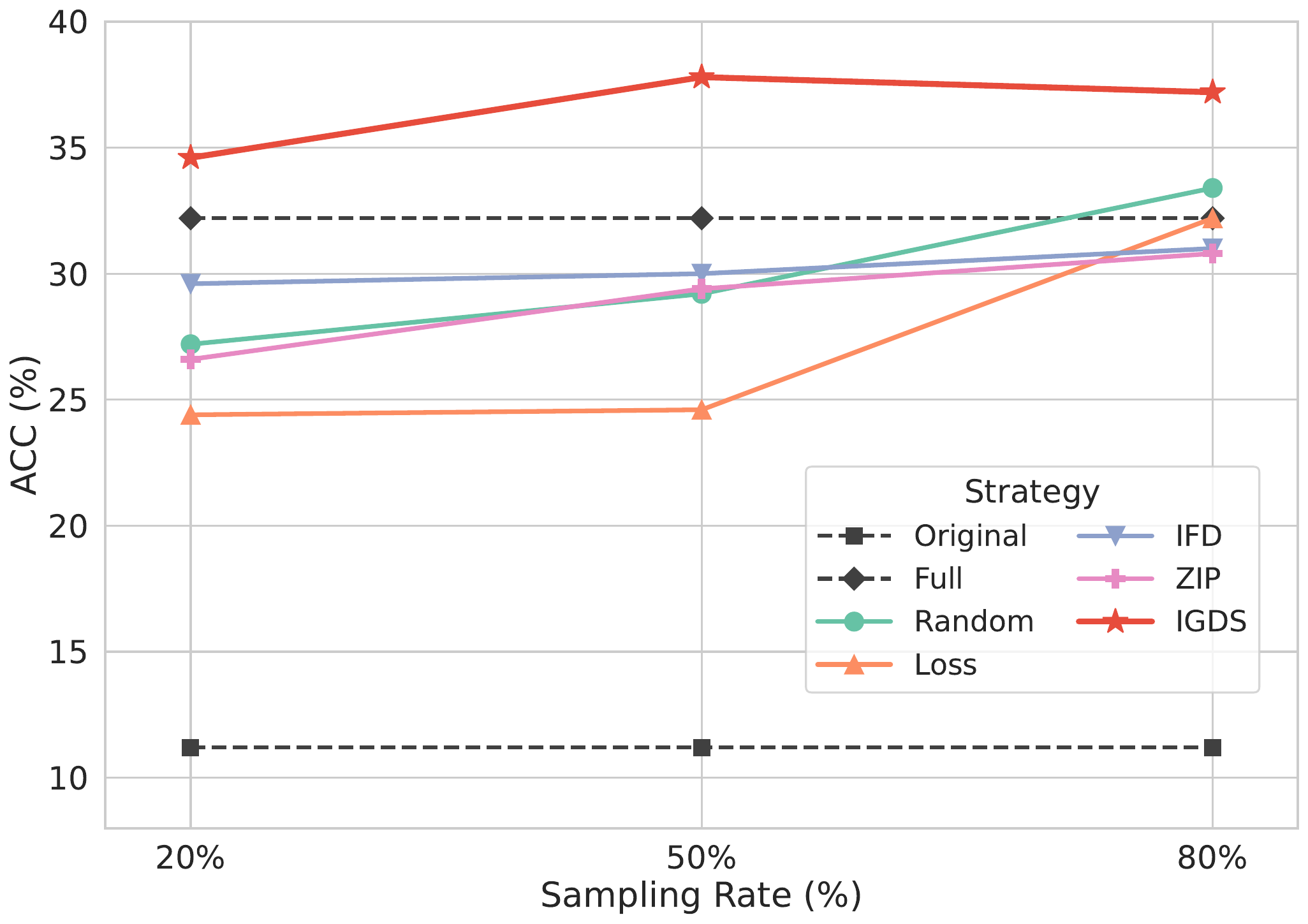} 
    \caption{Performance comparison of different data selection strategies under varying sampling rates (20\%, 50\%, 80\%) with \texttt{Gemma-2-2B} model.}
    \label{fig:sampling_robustness}
\end{figure}

\begin{table}[t!]
\centering
\small
\caption{Ablation study of the IGDS framework on the \texttt{Gemma-2-2B} model.}
\label{tab:ablation_study}
\resizebox{\linewidth}{!}{%
\begin{tabular}{lccc}
\toprule
\textbf{Method} & \textbf{Math} & \textbf{Sum} & \textbf{Trans} \\
\midrule
\textbf{Full IGDS (k=1)} & \textbf{37.8} & \textbf{0.452} & \textbf{68.14} \\
\midrule
\multicolumn{4}{l}{\textit{Ablation on Feature Identification Stage}} \\
\cmidrule(lr){1-1}
\quad w/o Frequency Recalling & 29.2 & 0.439 & 63.19 \\ 
\quad w/o Causal Filtering & 33.0 & 0.434 & 64.52 \\ 
\midrule
\multicolumn{4}{l}{\textit{Ablation on Data Selection Stage (Varying k)}} \\
\cmidrule(lr){1-1}
\quad $k=3$ & 37.2 & 0.447 & 66.81 \\
\quad $k=5$ & 31.8 & 0.435 & 62.83 \\
\bottomrule
\end{tabular}
}
\end{table}

To assess the robustness of our method under varying data budgets, we conducted experiments using sampling ratios of 20\%, 50\%, and 80\% of the full training set, specifically on the Math task with \texttt{Gemma-2-2B}. We benchmarked the proposed {IGDS} against all competitive baselines.

As illustrated in Figure~\ref{fig:sampling_robustness}, our {IGDS} exhibits overwhelming superiority across all tested sampling ratios. 
Notably, at every data budget, IGDS not only outperforms all other selection baselines by a significant margin but also consistently surpasses the performance of fine-tuning on the \textit{Full} dataset. These findings underscore the superior data efficiency of IGDS, demonstrating its capability to curate smaller, yet higher-utility subsets for effective and economical model training.

\subsection{Ablation Study}
To dissect the contributions of our framework's key components, we conducted an ablation study on the \texttt{Gemma-2-2B} model, with results presented in Table~\ref{tab:ablation_study}.
First, we first assess the impact of the feature identification pipeline. 
Replacing our frequency-based recalling with random selection (\textit{w/o Frequency Recalling}) leads to a sharp performance decline, underscoring the necessity of pre-screening the vast search space for relevant candidate features.
Similarly, removing the causal filtering step (\textit{w/o Causal Filtering}) and instead using all recalled features for scoring also significantly degrades performance. This confirms that causal validation is indispensable for distinguishing genuine task features from merely correlated noise.

Next, we examined sensitivity to $k$, the number of top features used for scoring. 
While our default setting ($k=1$) yields the best results, performance remains robust at $k=3$. 
However, increasing $k$ further to 5 results in a noticeable drop.
This observation suggests that a highly focused feature set is preferable, as including less impactful features likely introduces introduce noise and dilutes the selection quality.

\subsection{Time Cost of Data Selection}
\begin{table}[t!]
\centering
 
\caption{Time cost comparison of data selection strategies on \texttt{LLaMA-3.1-8B}. The raw time costs (in hours) are normalized relative to the \texttt{Loss} strategy (100\%).}\label{tab:efficiency}
\resizebox{\linewidth}{!}{%
\begin{tabular}{lccc}
\toprule
\textbf{Method} & \textbf{Math} & \textbf{Sum} & \textbf{Trans} \\
\midrule
Loss & 3.0 & 0.9 & 0.5 \\
IDF & 6.5\,(\textcolor{brown!40!red}{+167\%}) & 1.2\,(\textcolor{brown!40!red}{+33\%}) & 0.7\,(\textcolor{brown!40!red}{+40\%}) \\
ZIP & 1.6\,(\textcolor{teal!60!green}{-47\%}) & 0.4\,(\textcolor{teal!60!green}{-56\%}) & 0.2\,(\textcolor{teal!60!green}{-60\%}) \\ 
\textbf{IGDS} & 2.5\,(\textcolor{teal!60!green}{-17\%}) & 0.7\,(\textcolor{teal!60!green}{-22\%}) & 0.4\,(\textcolor{teal!60!green}{-20\%}) \\
\bottomrule
\end{tabular}
}
\end{table}

In addition to model performance, the computational efficiency of a data selection strategy is a critical determinant for its practical application. To evaluate this, we conducted a runtime analysis on the \texttt{LLaMA-3.1-8B} model by sampling 50\% of the data across three distinct tasks: Math, Summarization, and Translation. We measured the time cost of each strategy and normalized it relative to the \texttt{Loss} strategy, which serves as our 100\% baseline as it requires a single forward pass per sample.

As shown in Table~\ref{tab:efficiency}, the model-free \texttt{ZIP} strategy is naturally the most efficient (almost -60\%), whereas \texttt{IFD} incurs substantial overhead (+167\%) due to its need for extra inference in Math task. Our {IGDS} method demonstrates high efficiency, reducing the computational cost by approximately 20\% overhead. This is because its computation is integrated directly into the single forward pass required for gradient calculation. This result confirms that {IGDS} provides a practical solution for data selection without compromising on efficiency.

\section{Interpretability Analysis}
In this section, we analyze the structural distribution of the identified task-specific features and demonstrate their stability, confirming that they represent robust and intrinsic model properties.

\subsection{Distribution of Task-Specific Features}

\begin{figure}[t!]
    \centering
    \includegraphics[width=0.95\columnwidth]{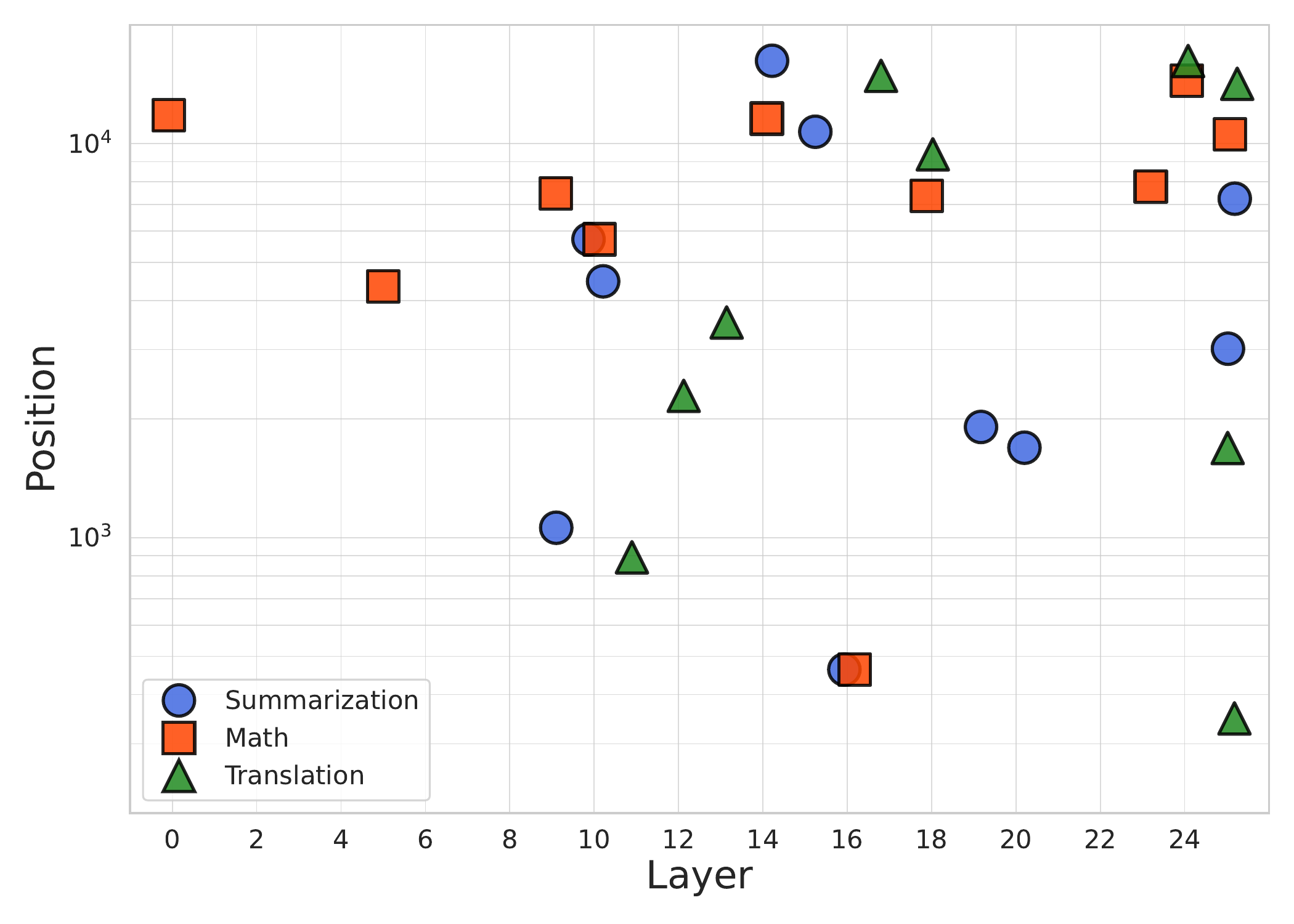} 
    \caption{Distribution of positive features across layers and positions for \texttt{Gemma-2-2B} model. Each point represents a feature, plotted by its layer (x-axis) and position (y-axis, log scale).}
    \label{fig:feature_distribution}
\end{figure}

To investigate where task-specific knowledge is encoded within the model, we visualize the distribution of identified features for Math, Summarization, and Translation tasks on the \texttt{Gemma-2-2B} model. As shown in Figure~\ref{fig:feature_distribution}, we plot each feature's layer against its position. Due to the vastness of the SAE's feature space, the position axis is presented on a logarithmic scale.
The results reveal distinct, task-dependent structural preferences. Specifically, Math features are widely dispersed across all layers, indicating that mathematical reasoning is a full-stack capability engaging the entire model depth. Conversely, Summarization and Translation features are concentrated in the middle-to-late layers, indicating a reliance on deep semantic processing. This differential distribution highlights that distinct cognitive abilities are localized in different structural regions of the model, underscoring the value of our fine-grained, task-specific analysis.


\subsection{Stability of Task-Specific Features}

\begin{table}[t!]
\centering
\caption{Stability of identified task features for Math on \texttt{Gemma-2-2B-it} using different prior datasets. The top-5 features with the highest positive impact are listed.}
\label{tab:feature_stability}
\begin{tabular}{c c c c}
\toprule
\textbf{Dataset} & \textbf{GSM8K} & \textbf{Math500} & \textbf{OpenR1} \\
\midrule
Top-1 & \cellcolor{red!20}\textbf{$F_{14,11575}$} & \cellcolor{red!20}\textbf{$F_{14,11575}$} & \cellcolor{red!20}\textbf{$F_{14,11575}$} \\
Top-2 & $F_{18,4651}$ & \cellcolor{blue!20}$F_{24,14448}$ & $F_{18,4651}$ \\
Top-3 & $F_{23,7783}$ & \cellcolor{gray!30}$F_{10,5717}$ & \cellcolor{blue!20}$F_{24,14448}$ \\
Top-4 & \cellcolor{blue!20}$F_{24,14448}$ & $F_{25,10550}$ & \cellcolor{gray!30}$F_{10,5717}$ \\
Top-5 & \cellcolor{gray!30}$F_{10,5717}$ & $F_{11,892}$ & $F_{5,4341}$ \\
\bottomrule
\end{tabular}
\end{table}

A crucial question for interpretability is whether identified features represent intrinsic model capabilities or are merely dataset-specific biases derived from the discovery source. To rigorously evaluate this stability, we performed independent feature identification runs on the Math task with \texttt{Gemma-2-2B-it}, utilizing three distinct datasets, GSM8K, Math500, and OpenR1, as separate identification sources. The results, presented in Table~\ref{tab:feature_stability}, reveal a remarkable degree of consistency across these diverse contexts.

Most strikingly, the top-ranked feature, \textbf{$F_{14,11575}$}, consistently emerges as the most impactful driver for mathematical reasoning, irrespective of the source dataset.
Furthermore, we observe a significant overlap within the top-5 ranks, with three specific features ($F_{14,11575}$, $F_{24,14448}$, and $F_{10,5717}$) persisting across all three settings.
This high degree of stability across varying data distributions provides compelling evidence that our framework is not simply finding patterns specific to one dataset's style or content.
Instead, it successfully identifies features that are fundamental and intrinsic to the model's core mechanism for mathematical reasoning.
\begin{figure*}[t!]
    \centering
    \begin{subfigure}[b]{0.32\textwidth}
        \centering
        \includegraphics[width=\textwidth]{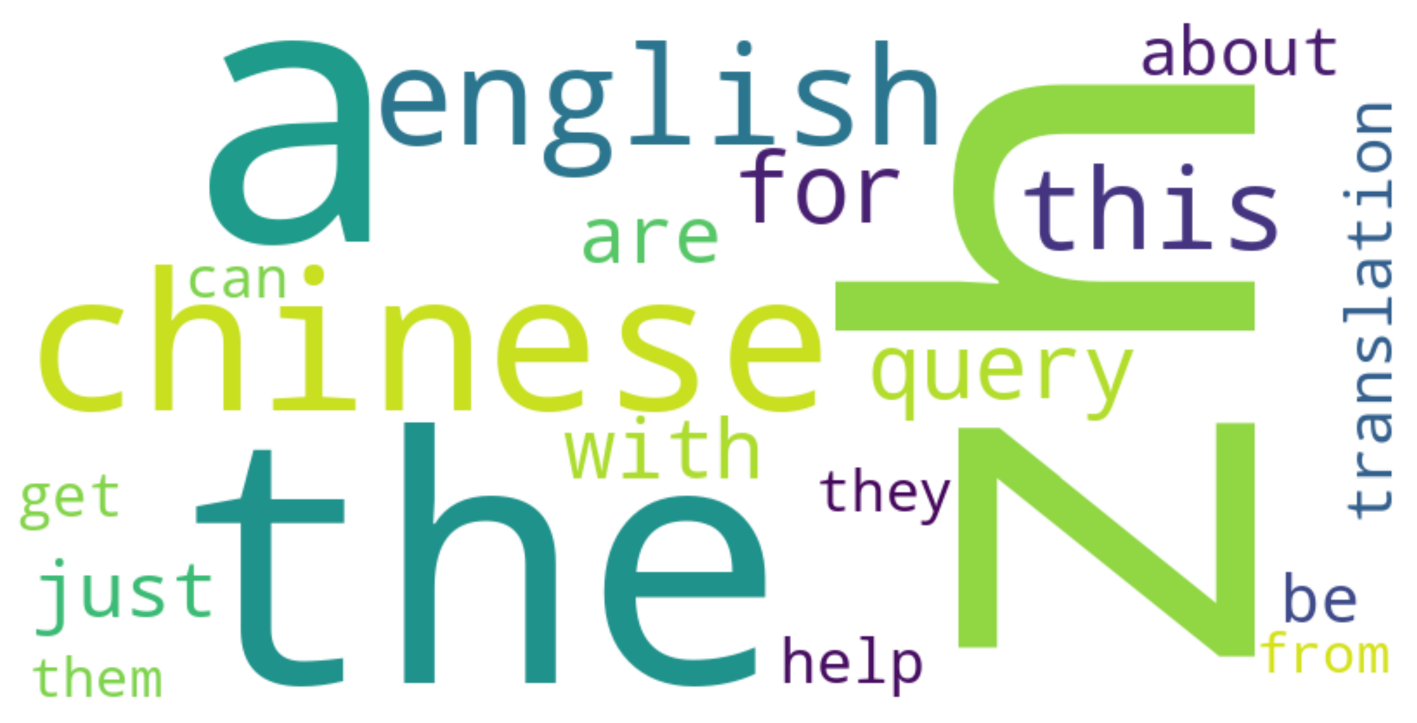}
        \caption{Translation Task}
        \label{fig:wordcloud_translation}
    \end{subfigure}
    \hfill
    \begin{subfigure}[b]{0.32\textwidth}
        \centering
        \includegraphics[width=\textwidth]{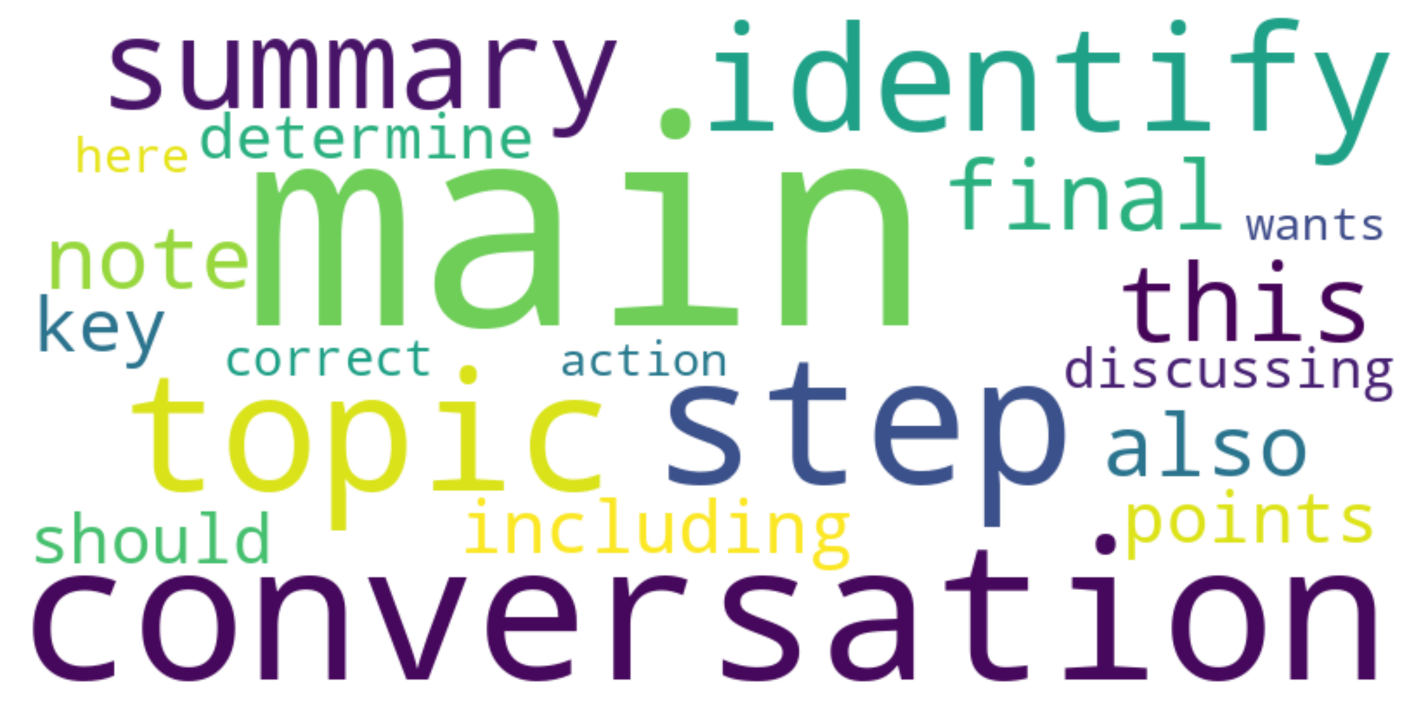}
        \caption{Summarization Task}
        \label{fig:wordcloud_summarization}
    \end{subfigure}
    \hfill
    \begin{subfigure}[b]{0.32\textwidth}
        \centering
        \includegraphics[width=\textwidth]{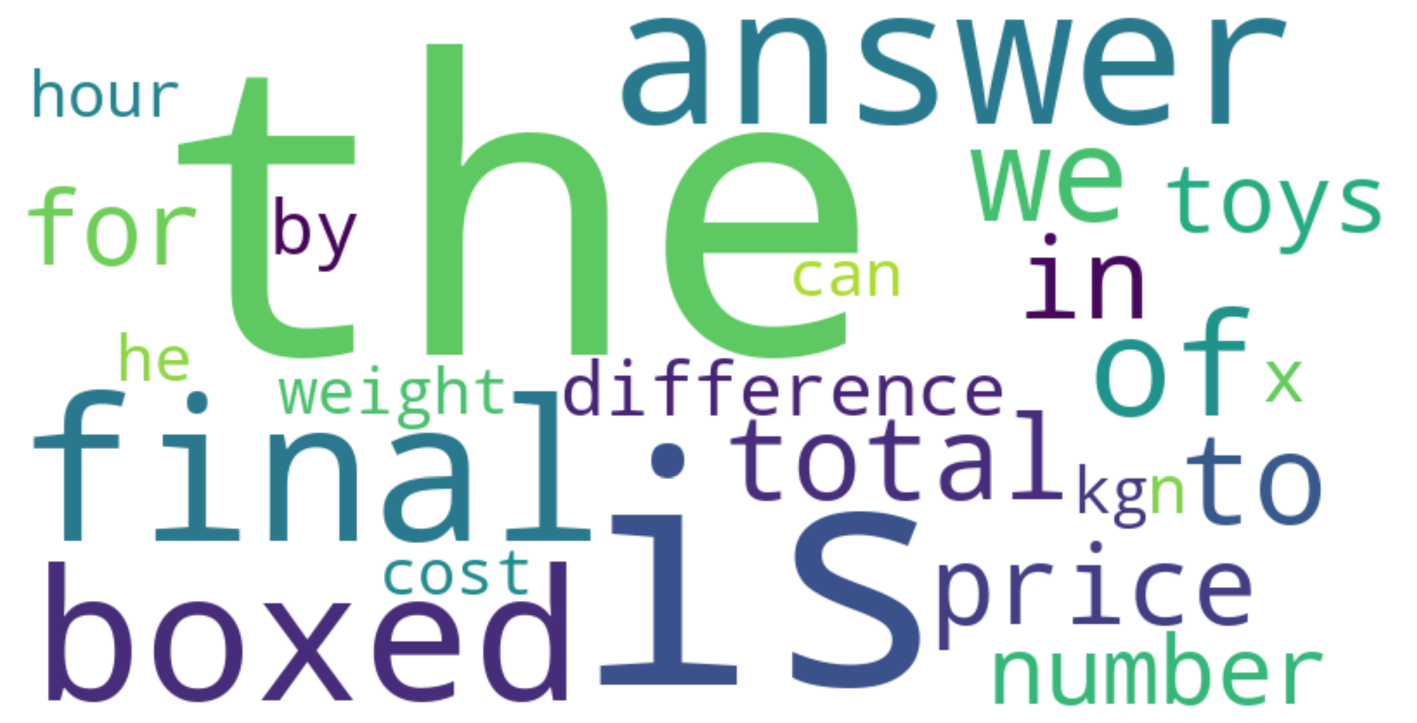}
        \caption{Math Task}
        \label{fig:wordcloud_math}
    \end{subfigure}
    \caption{Word clouds showing the terms with the most significant frequency increase after amplifying the top-ranked task feature for each task. The size of a word corresponds to the magnitude of its frequency change.}
    \label{fig:word_clouds}
\end{figure*}

\subsection{Correlation Analysis of Task Feature Identification}
\label{sssec:qualitative_analysis}


To better elucidate the semantic roles of the identified features, we qualitatively examine lexical shifts in the model's output when amplifying the top-ranked feature for each task. Figure~\ref{fig:word_clouds} visualizes the vocabulary exhibiting the most pronounced frequency increases, revealing a clear alignment with the underlying task semantics. For instance, amplifying the \textbf{Translation} feature consistently elevates language-specific terms such as ``english'' and ``chinese'' (Figure~\ref{fig:word_clouds}a). Likewise, strengthening the \textbf{Summarization} feature drives up the prominence of structural cues like ``summary'' and ``topic'' (Figure~\ref{fig:word_clouds}b), whereas activating the \textbf{Math} feature yields a marked increase in solution-oriented vocabulary such as ``answer'' and ``final'' (Figure~\ref{fig:word_clouds}c). This qualitative evidence appears consistent with our quantitative findings, supporting the notion that our data selection strategy is guided by features that encode meaningful, task-relevant concepts.

\section{Conclusion}
\label{sec:conclusion}

In this work, we have established and validated a new principle for LLM optimization: the most potent signals for data selection reside within the model's own internal causal mechanisms. We have presented Interpretability-Guided Data Selection (IGDS), a practical framework that operationalizes this principle by first identifying causally-validated task features, and subsequently selecting ``Feature-Resonant Data'' that maximally activates them. Our empirical results are compelling: IGDS not only consistently outperforms competitive baselines but can surpass full-dataset fine-tuning, achieving a remarkable \textbf{17.4\%} gain on the Math task using only half the data. Ultimately, IGDS pioneers a new class of optimization techniques that leverage interpretability, effectively bridging the gap between mechanistic understanding and practical model improvement.

\section*{Limitations}
\label{sec:limitations}

While IGDS presents a promising paradigm for data selection, we acknowledge that its performance is inherently linked to the quality of the underlying Sparse Autoencoders (SAEs). This dependency is empirically reflected in our results presented in Table~\ref{tab:main_results}, where model performance correlates strongly with the maturity of the available SAE ecosystems.


Specifically, for \texttt{Gemma-2-2B}, we utilized the high-quality, officially released Gemma-Scope SAEs. This resulted in the most significant performance improvements, with a particularly striking gain of $\mathbf{+17.4\%}$ on the Math task. In contrast, due to computational constraints, our custom SAE training for \texttt{Qwen3-8B} was restricted to a limited subset of layers. We hypothesize that this partial coverage of the model's layers is the primary reason for the more modest gains observed for Qwen3-8B, which were among the lowest across all tasks. 


This performance disparity underscores that the quality and comprehensiveness of the SAEs are critical factors that directly influence the efficacy of our method. To address this bottleneck and to contribute to the broader research community, we plan to train SAEs for the remaining layers of \texttt{Qwen3-8B} in our future work. We are committed to open-sourcing these artifacts upon completion to facilitate further research in this area.

\section*{Acknowledgments}
The present research was supported by the National Key Research and Development Program of China (Grant No. 2023YFE0116400). We would like to thank the anonymous reviewers for their insightful comments.

\bibliography{custom}

 \clearpage

\appendix

\section{Appendix}
\label{sec:appendix}

\subsection{Implementation Details}
\label{subsec:implementation}

This section provides additional details regarding our experimental setup to ensure the reproducibility of our work.

All experiments, including the training of our custom SAE for Qwen3 and all model fine-tuning runs, were conducted on a server equipped with 8 NVIDIA H100 GPUs. The implementation was based on the PyTorch framework, leveraging the \texttt{sae\_lens} and \texttt{LLaMA-Factory} libraries for efficient model handling and training. The use of this powerful hardware enabled the comprehensive evaluation of our framework across multiple models, tasks, and baselines.

\paragraph{Model Parameters} To ensure the reproducibility of our work, we provide detailed specifications for the Large Language Models (LLMs) and their corresponding Sparse Autoencoders (SAEs) in Table~\ref{tab:appendix_model_sae_params}. This includes architectural details of the base models and the configuration of the SAEs applied to them.

\paragraph{Supervised Fine-tuning}
All models were fully fine-tuned using the \texttt{LLaMA-Factory} framework on a high-performance computing cluster equipped with 8 NVIDIA H100 GPUs.
For each data subset selected by a given method, we trained the corresponding base model for one full epoch.
We used a consistent learning rate of $2 \times 10^{-5}$ and a global batch size of 64 across all experiments.
The computational efficiency of our IGDS framework in comparison to these baselines will be analyzed in a subsequent section.

\paragraph{Training SAEs}
\label{app:sae_training}
As no pre-trained SAE was publicly available for the \texttt{Qwen3-8B} model, we trained the classical SAE to facilitate our analysis. The training was implemented using the \texttt{sae\_lens} library (v6.6.0), employing a JumpReLU architecture on the residual stream outputs of layers 12 through 18. The dictionary size was set to 65,536, corresponding to an \textbf{16x expansion factor} over the model's hidden size. Key training hyperparameters included a context size of 512 tokens, a batch size of 2048, a learning rate of $5 \times 10^{-5}$, and a carefully tuned L1 coefficient for each layer to balance reconstruction loss and sparsity. The process was resource-intensive, requiring approximately \textbf{7 days} of computation on a single NVIDIA H100 GPU to train the SAE for one layer.

\subsection{Supplementary Results for Full ROUGE-1/2/L Metrics}
\begin{table}[t!]
\centering
 
\caption{Detailed parameters of the LLMs and their corresponding SAEs. Model names in the header are abbreviated for space.}
\label{tab:appendix_model_sae_params}
\resizebox{\linewidth}{!}{%
\begin{tabular}{@{}l ccc@{}}
\toprule
\textbf{Parameter} & \textbf{Gemma-2-2B} & \textbf{LLaMA-3.1-8B} & \textbf{Qwen3-8B} \\
\midrule
\multicolumn{4}{l}{\textit{LLM Parameters}} \\
\midrule
\quad Total Layers & 26 & 32 & 36 \\
\quad Hidden Size ($d_{\text{model}}$) & 2,304 & 4,096 & 4,096 \\
\midrule
\multicolumn{4}{l}{\textit{SAE Parameters}} \\
\midrule
\quad SAE Source & Gemma-scope & Llama-scope & Self-trained \\
\quad Dict. Size ($d_{\text{sae}}$) & 18,432 & 32,768 & 65,536 \\
\quad Exp. Factor & 8x & 8x & 16x \\
\quad Applied Layers & [0--25] & [0--31] & [12--18] \\
\bottomrule
\end{tabular}
}
\end{table}

\label{full-rouge}
Table~\ref{tab:my-table-full-rouge} details the ROUGE-1, ROUGE-2, and ROUGE-L metrics for the summarization task, supplementing the main results in Table~\ref{tab:main_results}. Across all three models (\texttt{Gemma-2-2B}, \texttt{Llama-3.1-8B}, and \texttt{Qwen3-8B-Base}), our {IGDS} method consistently outperforms other data selection baselines (Random, Loss, IFD, and ZIP). Notably, IGDS frequently achieves superior performance compared to training on the {Full} dataset---securing the highest scores on all metrics for Qwen3-8B-Base and leading in most metrics for the other models---demonstrating its effectiveness in selecting high-quality samples for text generation.
\begin{table}[t!]
	\centering
 \caption{Detailed ROUGE-1/2/L results for the summarization task across different models.}
	\label{tab:my-table-full-rouge}
    \resizebox{\linewidth}{!}{%
	\begin{tabular}{llccc}
 \toprule
		               \textbf{Model}        &\textbf{Method}& \textbf{Rouge-1} & \textbf{Rouge-2} & \textbf{Rouge-L} \\
                         \midrule
		\multirow{7}{*}{\textbf{Gemma-2-2B}}& Base & 0.2203 & 0.0731 & 0.1698 \\
 & Full & 0.4501 & 0.2038 & \textbf{0.3630} \\
 & Random & 0.4394 & 0.1954 & 0.3494 \\
 & Loss & 0.4480 & 0.2003 & 0.3589 \\
 & IFD & 0.4400 & 0.1999 & 0.3565 \\
 & ZIP & 0.4491 & 0.1932 & 0.3607 \\
 & \textbf{IGDS} & \textbf{0.4522} & \textbf{0.2089} & 0.3627 \\ \midrule
		\multirow{7}{*}{\textbf{Llama-3.1-8B}}& Base & 0.0107 & 0.0039 & 0.0084 \\
 & Full & 0.2600 & \textbf{0.1230} & 0.2145 \\
 & Random & 0.2542 & 0.1174 & 0.2155 \\
 & Loss & 0.2539 & 0.1158 & 0.2148 \\
 & IFD & 0.2578 & 0.1166 & 0.2149 \\
 & ZIP & 0.2451 & 0.1080 & 0.2109 \\
 & \textbf{IGDS} & \textbf{0.2606} &0.1187 & \textbf{0.2198} \\ \midrule
		\multirow{7}{*}{\textbf{Qwen3-8B-Base}}  & Base & 0.3118 & 0.1243 & 0.2515 \\
 &Full & 0.4892 & 0.2247 & 0.3954 \\
 & Random & 0.4823 & 0.2185 & 0.3892 \\
 & Loss & 0.4741 & 0.2126 & 0.3817 \\
 & IFD & 0.4884 & 0.2232 & 0.3939 \\
 & ZIP & 0.4849 & 0.2198 & 0.3916 \\
 & \textbf{IGDS} & \textbf{0.4901} & \textbf{0.2274} & \textbf{0.3968} \\\bottomrule
	\end{tabular}
	}
\end{table}

\begin{table*}[t!]
 
\centering

\caption{Performance results of general capabilities on MMLU and TruthfulQA benchmarks.}\label{tablegemma}

\begin{tabular}{lcccccc}
\toprule
\multirow{2}{*}{} & \multicolumn{2}{c}{Math} & \multicolumn{2}{c}{Sum} & \multicolumn{2}{c}{Trans} \\ \cmidrule{2-7} 
       & MMLU & TruthfulQA & MMLU & TruthfulQA & MMLU & TruthfulQA \\ \midrule
{Base} & 54.12\% & 36.23\% & 54.12\% & 36.23\% & 54.12\% & 36.23\% \\\midrule

{IGDS} & 53.25\% & 36.46\% & 53.18\% & 33.57\% & 54.05\% & 35.88\% \\

{Full} & 53.75\% & 36.00\% & 52.67\% & 34.00\% & 53.12\% & 34.35\% \\

{Random} & 53.77\% & 35.63\% & 53.23\% & 31.74\% & 52.98\% & 36.12\% \\

{ZIP} & 53.39\% & 36.41\% & 51.84\% & 34.12\% & 53.67\% & 33.91\% \\

{IFD} & 53.21\% & 38.33\% & 52.45\% & 35.06\% & 51.33\% & 34.72\% \\

{Loss} & 53.84\% & 37.63\% & 54.01\% & 33.44\% & 52.19\% & 35.21\%\\
\bottomrule
\end{tabular}

\end{table*}
\subsection{Prompt Template for Task Feature Identification}

\label{app:prompt_templates}

Table~\ref{tab:prompts} presents the detailed prompt templates employed across three distinct phases of our experiments: Task Feature Identification (Stage 1), Supervised Fine-tuning, and Evaluation. These templates cover the three downstream tasks: Mathematical Reasoning (Math), Text Summarization (Sum), and Machine Translation (Trans).

In the table, text enclosed in curly braces and formatted in italics (e.g., \textit{\{Question\}}, \textit{\{Solution\}}) represents data-specific placeholders. These are replaced by the actual content of the corresponding samples from the dataset during processing.

For the \textbf{Task Feature Identification} stage, we specifically mark the position used for feature extraction. The colon symbol highlighted with a yellow background and red font (\colorbox{yellow}{\textcolor{red}{\textbf{:}}}) indicates the exact token index where the model's internal hidden states (activations) are computed and extracted. This position serves as the final representation of the input context before the generation of the target sequence begins.

For the \textbf{Evaluation} stage, we employ a 4-shot setting for the Math task to ensure stable reasoning performance, while the Summarization and Translation tasks are evaluated in a zero-shot setting to test the model's direct instruction-following capabilities.

\begin{table*}[t] %
\centering
\small
 \caption{Prompt used in Different stages for different tasks.}
 \label{tab:prompts}
 
 \begin{tabularx}{\textwidth}{ >{\raggedright\arraybackslash}p{1cm} c X } 
 \toprule
	\textbf{Stage} & \textbf{Task}  & \textbf{Prompt}     \\ 
 \midrule
 
\multirow{3}{=}[-0.5em]{Task Feature Identification}
		& Math  &    \textit{\{Question\}} \textbackslash n
Please reason step by step, and put your final answer within \textbackslash boxed\{\}. \textbackslash n Solution\colorbox{yellow}{\textcolor{red}{\textbf{:}}} \textit{\{Solution\}} \textbackslash n The final answer is \textbackslash boxed\{\textit{\{Golden Answer\}}\}.  \\
		& Sum   & Use a sentence to summarize this following text: \textbackslash n \textit{\{Dialogue\}} \textbackslash n Summarization\colorbox{yellow}{\textcolor{red}{\textbf{:}}} \textit{\{Summary\}} \\
		& Trans &  Please translate the following text into \textit{\{Target\_language\}}.\textbackslash n Text:\textit{\{Source\_text\}} \textbackslash n Translation\colorbox{yellow}{\textcolor{red}{\textbf{:}}} \textit{\{Target\_text\}}        \\ 
  \midrule 
  
\multirow{3}{=}[-2.5em]{Supervised Fine-tuning}
		& Math  &Instruction: \textit{\{Question\}} \textbackslash n
Please reason step by step, and put your final answer within \textbackslash boxed\{\}. \textbackslash n \newline Input: \newline Output: Solution: \textit{\{Solution\}} \textbackslash n The final answer is \textbackslash boxed\{\textit{\{Golden Answer\}}\}.\\
		& Sum   &Instruction: Use a sentence to summarize this following text.\newline Input: \textit{\{Dialogue\}}\newline Output: \textit{\{Summary\}}          \\
		& Trans &Instruction: Please translate the following text into \textit{\{Target\_language\}}.\newline Input: \textit{\{Source\_text\}}\newline Output: \textit{\{Target\_text\}}         \\
  \midrule
  
  \multirow{3}{=}[-0.5em]{Evaluation}
		& Math  & Please reason step by step, and put your final answer within \textbackslash boxed\{\} as the following format. \textbackslash n Here are some examples: \textit{\{4-Shots\}}  \textbackslash n
            \#\#\# Problem:\textbackslash n \textit{\{Question\}}\textbackslash n\#\#\# Solution:           \\
		& Sum   &   Use a sentence to summarize this following text: \textbackslash n \textit{\{Dialogue\}} \textbackslash n Summarization:       \\
		& Trans & Please translate the following text into \textit{\{Target\_language\}}.\textbackslash n Text:\textit{\{Source\_text\}} \textbackslash n Translation:         \\
  		
  \bottomrule\end{tabularx}

\end{table*}

\subsection{Detailed Topology of Task-Specific Features}
\label{sec:feature_topology}

\begin{figure*}[htb]
    \centering
    \begin{subfigure}[b]{\textwidth}
        \centering
        \includegraphics[width=\textwidth]{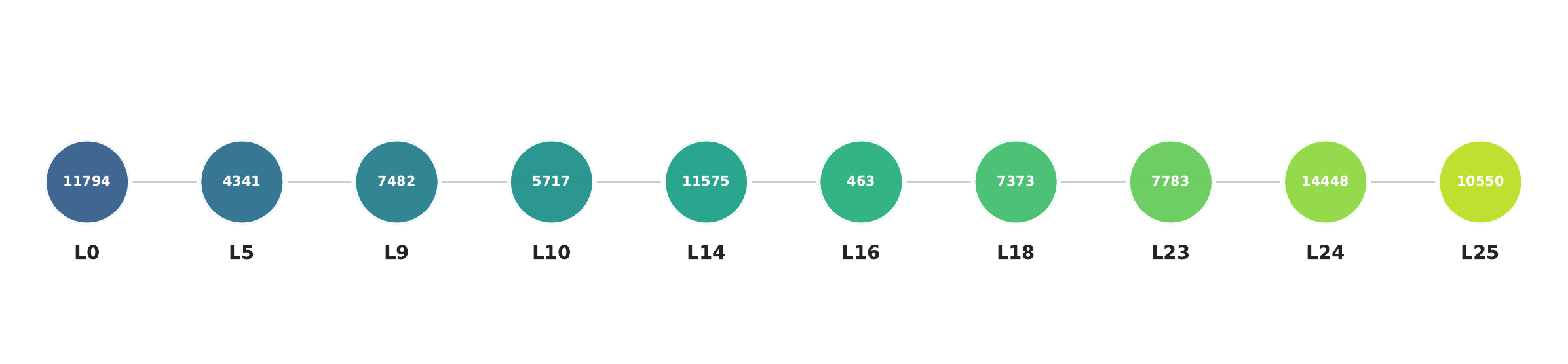}
        \caption{Topological signature of features for Math Task, which exhibits a globally distributed topology. Features are activated across the entire depth (L0-L25).}
        \label{fig:flow_math}
    \end{subfigure}

    \begin{subfigure}[b]{\textwidth}
        \centering
        \includegraphics[width=\textwidth]{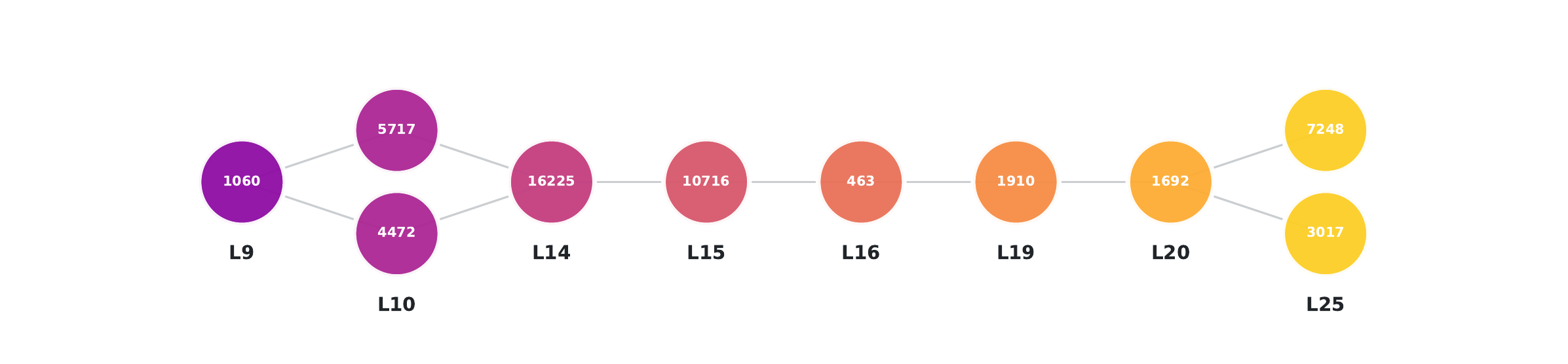}
        \caption{Topological signature of features for Summarization Task, in which features are absent in early layers and emerge primarily from L9 onwards.}
        \label{fig:flow_sum}
    \end{subfigure}

    \begin{subfigure}[b]{\textwidth}
        \centering
        \includegraphics[width=\textwidth]{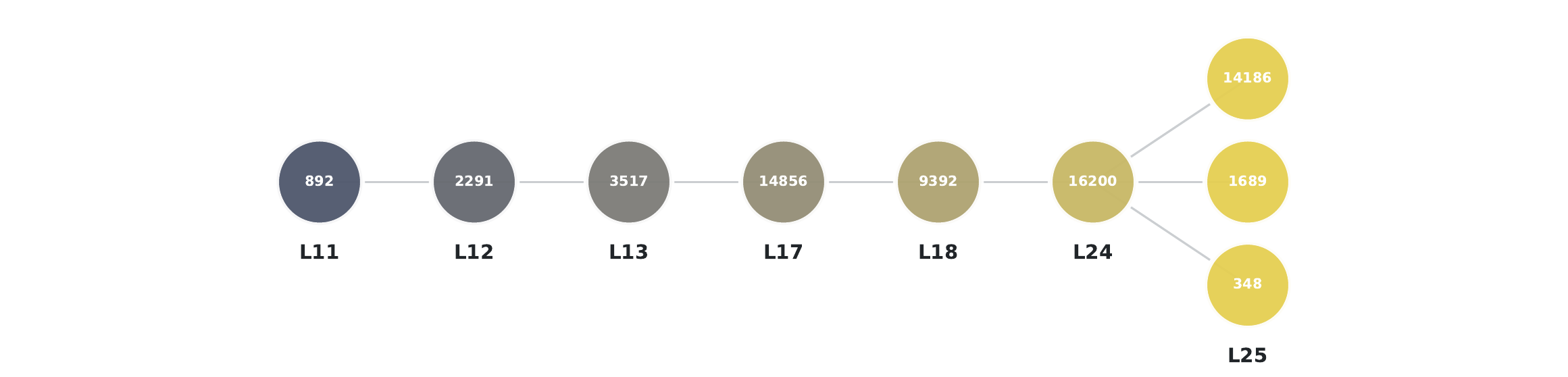}
        \caption{Topological signature of features for Translation Task, which shows a sparse, localized activation pattern concentrated in middle (L11-L18) and late layers.}
        \label{fig:flow_translation}
    \end{subfigure}

    \caption{Fine-Grained Topology of Task-Specific Features in Gemma-2-2B.
    This figure serves as a microscopic supplement to the macroscopic distribution shown in Figure~\ref{fig:feature_distribution}.}
    \label{fig:gemma_feature_flow}
\end{figure*}

While Figure~\ref{fig:feature_distribution} provides a macroscopic view of the structural distribution, illustrating \textit{where} task-specific knowledge is generally concentrated, it is equally crucial to identify \textit{what} these features specifically are to confirm they represent stable, intrinsic model properties. To this end, we visualize the fine-grained topological signatures of the identified \textbf{Task-Specific Features} for the Math, Summarization, and Translation tasks in Figure~\ref{fig:gemma_feature_flow}.


    


\subsection{Evaluation of General Capabilities}

To assess the impact of various methods on fundamental performance, we evaluated \texttt{Gemma-2-2B} and its different fine-tuning version across MMLU~\citep{hendryckstest2021} and TruthfulQA~\citep{lin-etal-2022-truthfulqa} benchmarks (Table~\ref{tablegemma}). The results demonstrate that SFT does not adversely affect general capabilities: across all methods (IGDS, Full, Random, ZIP, IFD, Loss), performance fluctuations remain within a narrow margin, typically between $-3\%$ and $+1\%$ relative to the Base model. The models maintain high accuracy in reasoning and truthfulness across Math, Summarization, and Translation tasks, showing no signs of catastrophic forgetting. The result confirms that our fine-tuning strategies successfully preserve the core knowledge and logical proficiency of the foundation models while achieving task alignment.

\end{document}